\documentclass{article}
\usepackage[numbers]{natbib}
\usepackage[pdftex]{graphics}
\usepackage{color}
\usepackage{kotex}
\usepackage{graphbox}
\usepackage{caption}
\usepackage{subcaption}

\usepackage[preprint]{nips_2018}

\usepackage[utf8]{inputenc} % allow utf-8 input
\usepackage[T1]{fontenc}    % use 8-bit T1 fonts
\usepackage{hyperref}       % hyperlinks
\usepackage{url}            % simple URL typesetting
\usepackage{booktabs}       % professional-quality tables
\usepackage{amsfonts}       % blackboard math symbols
\usepackage{nicefrac}       % compact symbols for 1/2, etc.
\usepackage{microtype}      % microtypography

\usepackage{xcolor}

\title{Doubly Nested Network \\ for Resource-Efficient Inference}
\author{
  Jaehong Kim
  \And
  Sungeun Hong
  \And
  Yongseok Choi
  \And
  Jiwon Kim
  \And
  \normalfont{SK T-Brain} \\
  \texttt{\{xhark, csehong, yschoi, jk\} @sktbrain.com}
}

\begin{document}

\maketitle
\begin{abstract}
%     주어진 문제와 유사한 대규모 데이터셋으로부터 학습한 정보는 문제해결이 큰 도움을 주지만, 그걸 위해 Scratch부터 학습하는 것은 시간이 오래걸린다. 그래서 일반적으로는 공개된 Pre-trained 모델을 가져와 사용한다. 하지만 공개된 Pre-trained 모델의 종류는 제한적이라, 내가 원하는 Budget Constraint에 맞는 모델을 찾는것은 어려운 일이다. 보통은, 성능좋은 Network로 부터 Prunning으로 Budget를 맞추고 Finetune을 통해 성능을 보정한다. 하지만 이러한 방법은 기존 Network가 학습한 Data가 필요한 문제점이 있다. 우리는 이러한 Data가 없는 경우에도 (Finetune없이) 동작하는 Sliceable한 구조를 제안한다.

We propose doubly nested network(DNNet) where all neurons represent their own sub-models that solve the same task. Every sub-model is nested both layer-wise and channel-wise. While nesting sub-models layer-wise is straight-forward with deep-supervision as proposed in \cite{xie2015holistically}, channel-wise nesting has not been explored in the literature to our best knowledge. Channel-wise nesting is non-trivial as neurons between consecutive layers are all connected to each other. In this work, we introduce a technique to solve this problem by sorting channels topologically and connecting neurons accordingly. For the purpose, channel-causal convolutions are used. Slicing doubly nested network gives a working sub-network. The most notable application of our proposed network structure with slicing operation is resource-efficient inference. At test time, computing resources such as time and memory available for running the prediction algorithm can significantly vary across devices and applications. Given a budget constraint, we can slice the network accordingly and use a sub-model for inference within budget, requiring no additional computation such as training or fine-tuning after deployment. We demonstrate the effectiveness of our approach in several practical scenarios of utilizing available resource efficiently. 

%As we cannot prepare combinatorially many different models for all situations, designing a neural network architecture that is resource-adaptive at test time (that exactly uses the given time and space) is very important. In this paper, we first propose a network structure and learning method to deliver a model equipped with many different smaller models with varying recourse demands in it. Our proposed network structure ... blah blah ... . To train this, we use ... blah ... blah .. scheme. Experiments on ... blah .. tasks demonstrate that our proposed method competitively perform considering that we use the (almost) exact amount of available budgets. 

\end{abstract}

\section{Introduction}

\begin{figure*}[htbp]
	\begin{subfigure}{1\textwidth}
	\centering		
    \includegraphics[width=0.915\linewidth]{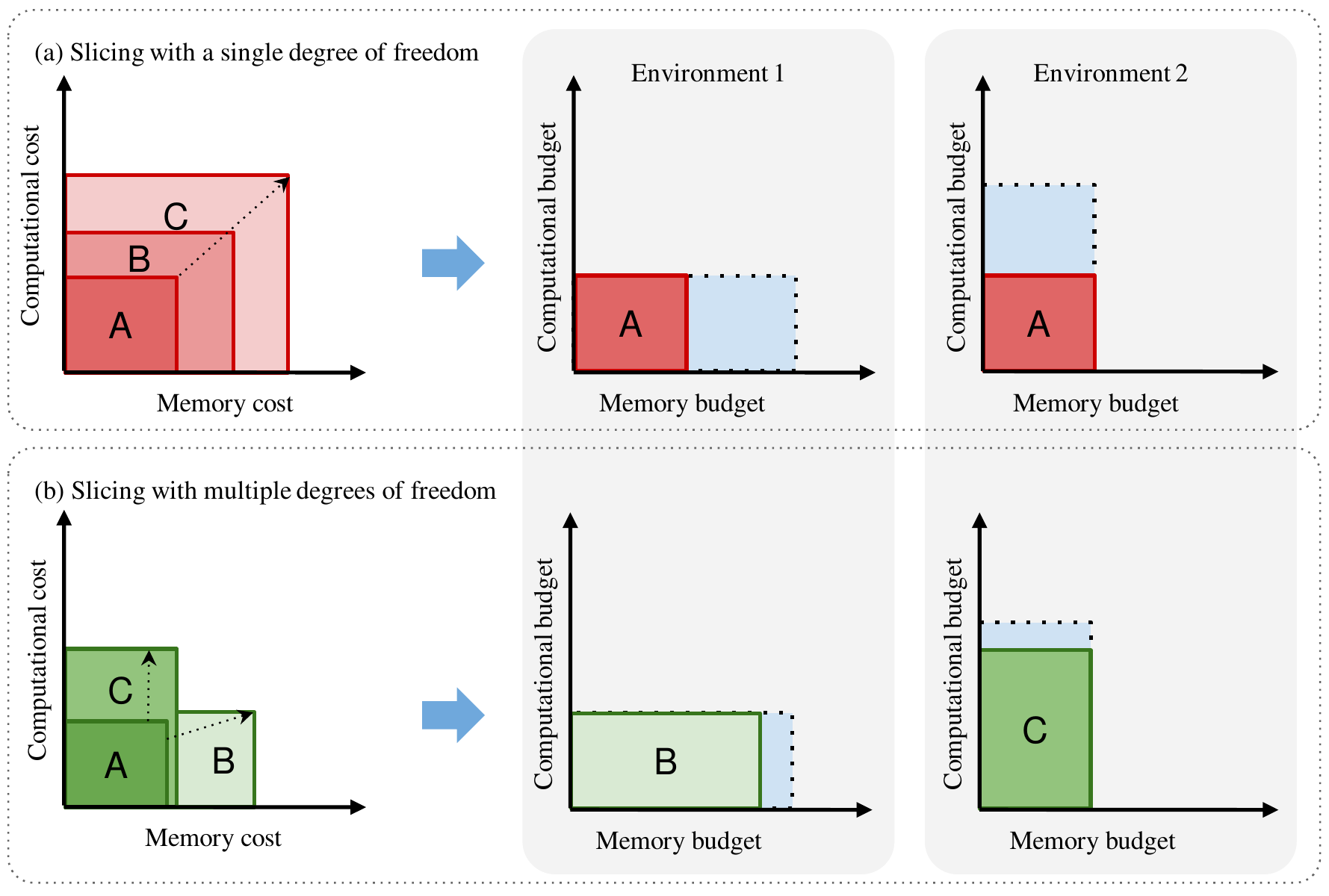}
% 		\caption{}
	\end{subfigure}
	\caption{
    (a) Sliceable models generated with a single control parameter (e.g., width only or depth only) have tight correlation between their computational and memory costs. These model might suffer from resource underutilization in various environments with different computational and memory budgets, which result in underutilization of memory capacity in computation-hungry `environment 1' or computational capacity in memory-hungry `environment 2' (b) on the other hand, having more degrees of freedom (e.g., both width and depth) allows to generate various sliced models with different computational and memory costs and achieve higher resource utilization.
    }
	\label{fig:our_adv_concept}
\end{figure*}

%Modern Deep Learning Application 들은 공개된 pretrained model을 가져와 쓰는 경우가 대부분이다. Image Domain에서는 ImageNet을 학습한 CNN feature extractor나, Language Domain 에서는 Wiki로 학습된 Word vectors embedding등이 있다. Image의 경우 학습된 CNN의 파라매터를 가져다 그냥 사용하거나, 혹은 일부 parameter를 finetune하여 사용한다. 이 경우 Deploy할 때에는 속도등 Budget Constraint가 있으면 Prunning이나 Quantization등으로 경량화하고 다시 네트워크를 Finetune하는 과정이 필요하다. Finetune을 위해서는 네트워크가 학습했던 Data를 필요로 하는데, 성능좋은 public feature extractor의 경우 매우 거대한 Dataset (like ImageNet)으로 학습되어 있어 Finetune이 어렵게된다. (From Scratch training 보다는 낫지만.)

Deep neural network (DNN) based methods have shown remarkable performance over a wide variety of fields \cite{szegedy2015going, girshick2014rich, kim2017learning, shin2017continual, chen2018deeplab}. Some of notable deep learning architectures such as ResNet \cite{DBLP:conf/cvpr/HeZRS16} or DenseNet \cite{DBLP:conf/cvpr/HuangLMW17} stack layers very deep or very wide and they result in very high performance. Most methods that have performed well in competitions or benchmark datasets have elaborate architecture designs typically requiring too high capacity for to be deployed in practice. Computing power is never free and therefore applying sophisticated but resource-intensive models directly to real environments with time or memory budget constraints is impractical especially in the context of mobile devices or embedded systems. 

%\cite{DBLP:journals/corr/HuangCLWMW17:msdnet_iclr18}.

To address this issue, techniques have been proposed which can improve the memory or time efficiency by eliminating the redundancy of the existing network model.
Among them, pruning is the most widely used technique and this approach usually cut back connections between layers \cite{DBLP:conf/nips/AlvarezS16, DBLP:conf/eccv/ZhouAP16}, the number of layers \cite{DBLP:conf/nips/WenWWCL16}, or the number of channels \cite{DBLP:conf/iccv/HeZS17}. Another strategy involves methods using low-rank approximation of weight tensors by minimizing the difference between the original network and the reconstructed reduced network \cite{comp_mobile_iclr_2015:DBLP:conf/iclr/KimPYCYS15, DBLP:journals/pami/ZhangZHS16}.
In addition, there has been a growing interest in knowledge distillation based methods that build a compact network  from a pre-trained large network while maintaining  the performance of a large network \cite{knowledge_distillation:DBLP:journals/corr/HintonVD15}. 
Overall, in order to reduce the redundancy of the original model, these techniques have in common that the compact model must be re-trained each time a new specification or constraint is required. 

Although the above techniques have shown the applicability of the learned model in the low capacity environment, there is still one issue that  has been rarely addressed  yet.
Considering the practical application of techniques to reduce model redundancy, a relatively large original network needs to be applied to numerous devices with  different specifications  or budget constraints rather than being applied once to a specific device.
Unfortunately, conventional techniques re-train the original network again every time the environment specification changes in terms of memory or time, so it is inefficient and infeasible to apply them when there are diverse target environments.

%
%We formulate this issue as a problem of automatically learning a neural network architecture under budgeted con- straints. To tackle this problem, we propose a budgeted learning approach that integrates a maximum cost directly in the learning objective function. The main originality of our approach with respect to state-of-the-art is the fact that it can be used with any type of costs, existing methods be- ing usually specific to particular constraints like inference speed or memory consumption – see Section 5 for a com- plete review of state-of-the-art. In our case, we investigate the ability of our method to deal with three different costs: (i) the computation cost reflecting the inference speed of the resulting model, (ii) the memory consumption cost that measures the final size of the model, and the (iii) distributed computation cost that measures the inference speed when computations are distributed over multiple machines or pro- cessors.

The most challenging scenario with regards to resource budget is when both time and memory available for inference dynamically change as shown in Fig.~\ref{fig:our_adv_concept}. If our deployed model is supposed to use time or memory exceeding the limits, we need a mechanism for the model to be adapted to utilize available resource only while maintaining its original performance as high as possible. It is widely known that the number of layers is one of dominant factors determining inference time and the number of channels dominantly affects maximum required memory. While there can be several directions to tackle the challenge of reducing time and memory usage dynamically, a baseline is to prepare many different models with varying resource requirements. This requires training models many times and storing all of them consumes too much of space. One can expect that models with slight differences in the number of layers or channels mostly share similarities. It is ideal for all models to share as many parameters as possible to reduce training time and memory to store them. 

To resolve the aforementioned issues, we propose a neural network architecture called doubly nested network(DNNet) where all models share parameters maximally by nesting all-in-one. Nesting models layer-wise is not a new concept. It is proposed in \cite{xie2015holistically}, where each layer has its side output and deeply supervised using \cite{lee2015deeply}. 

While nesting models layer-wise is relatively straight-forward, nesting them channel-wise is challenging. One of the reasons for the difficulty is that all channels between two consecutive layers are fully connected and slicing them seems unnatural. To resolve the issue, we propose to sort channels topologically so that some channels are relatively more important than the others. This can be implemented with channel-causal convolution. Since information flows from lower indexed channels to higher index channels and not vice versa, prediction with only lowest channels still works. 

Slicing DNNet does not need further re-training to satisfy each new specification or budget constraint. Allowing both vertical and horizontal slicing, inherited from doubly nested structure, is very important since there can be scenarios where time (usually correlated with computational cost) is the limiting factor or memory is the limit factor as illustrated in Fig.~\ref{fig:our_adv_concept}.

\section{Related Works}

Along with the growth of deep neural networks, considerable efforts have been devoted to reduce the redundancy of original model or propose the resource efficient model for applications in environments that require low computation memories as well as fast inference speed. Early investigation of these approaches focus on network pruning, which is a conventional way to reduce the network time and space complexities. Han et al. \cite{DBLP:conf/nips/HanPTD15:nips_2015,DBLP:conf/iclr/HanMD16:iclr_2016} proposed an approach that iteratively prune and retrain the network using regularization \cite{DBLP:conf/nips/HanPTD15:nips_2015} and also showed that more reduction could be achieved by combining trained quantization techniques \cite{DBLP:conf/iclr/HanMD16:iclr_2016}. Structured or coarse-grained sparsity has been also studied to achieve actual speed--up in common parallel processing environment including channel-level pruning \cite{DBLP:conf/iccv/HeZS17, DBLP:conf/nips/WenWWCL16} and layer-level pruning \cite{Wen:2016:LSS:3157096.3157329}. 

Although the above techniques have shown the applicability of the learned model in the low capacity environment, they commonly require retraining or fine-tuning to build the model to meet continuously changing budget constraints. To address this issue, some pioneering approaches have been proposed in terms of inference speed or memory footprint. Runtime neural pruning (RNP) \cite{DBLP:conf/nips/LinRLZ17:rnp_nips2017} is similar to our method in that the both can dynamically adjust the channel size per layer at run time. However, unlike our model, which can be deployed after slicing according to the budget constraint, RNP preserves the full architecture of the original network and conducts pruning adaptively, which is beneficial only in terms of time and not in terms of parameter memory. In considering a single network that can maximizes accuracy on various devices which have different computational constraints, Multi-Scale DenseNet (MSDNet) \cite{DBLP:journals/corr/HuangCLWMW17:msdnet_iclr18} is closely related to our approach.
They produce features of all scales (fine-to-coarse) by the multi-scale architecture, which facilitates good classification early on but also extracts low-level features that only become useful after several more layers of processing. To slice the original network along depth axis (i.e., layer-wise), they used multiple early exit classifiers in the middle of the layers. In contrast, we have devised a slicing technique in the width axis (i.e., channel-wise) which has not been well investigated previously, and ultimately propose an architecture capable of slicing at a higher degree of freedom, taking both width and depth into consideration. 

The optimal architecture selection for a given task has been one of the long standing challenges in designing neural network-based algorithms. A few previous studies have addressed this problem by giving more flexibility to the selection of width and depth of the network. In \cite{DBLP:conf/nips/SaxenaV16:neural_fabrics_nips2016}, a three-dimensional trellis-like architecture, called ``convolutional neural fabrics'' was introduced to tackle this challenge by training a large network which is able to embed multiple architectures with different architectural hyper-parameters (e.g., the number of stride at each layer, the number of channels at each layer) conceptually. This architecture has been adopted as a baseline in one algorithm about discovering cost-constrained optimal architectures \cite{DBLP:journals/corr/VeniatD17:budgeted_super_networks}. Our work  benefits from a large degree of freedom in choosing the optimal neural network architectures, but, differs from these works in that they do not consider extraction of partial models with various computation and memory costs from a large model without retraining or fine-tuning.

\section{Proposed method}

\subsection{Network architecture of DNNet}

%As can be seen in the figure, there are classifiers for each channel or layer.
%Their role is to allow for early exit when inference or to truncate the original model according to the budget constraints of the model we would like to deploy.

\begin{figure}
	\begin{subfigure}[c]{\linewidth}
		\centering
		\includegraphics[width=0.95\linewidth]{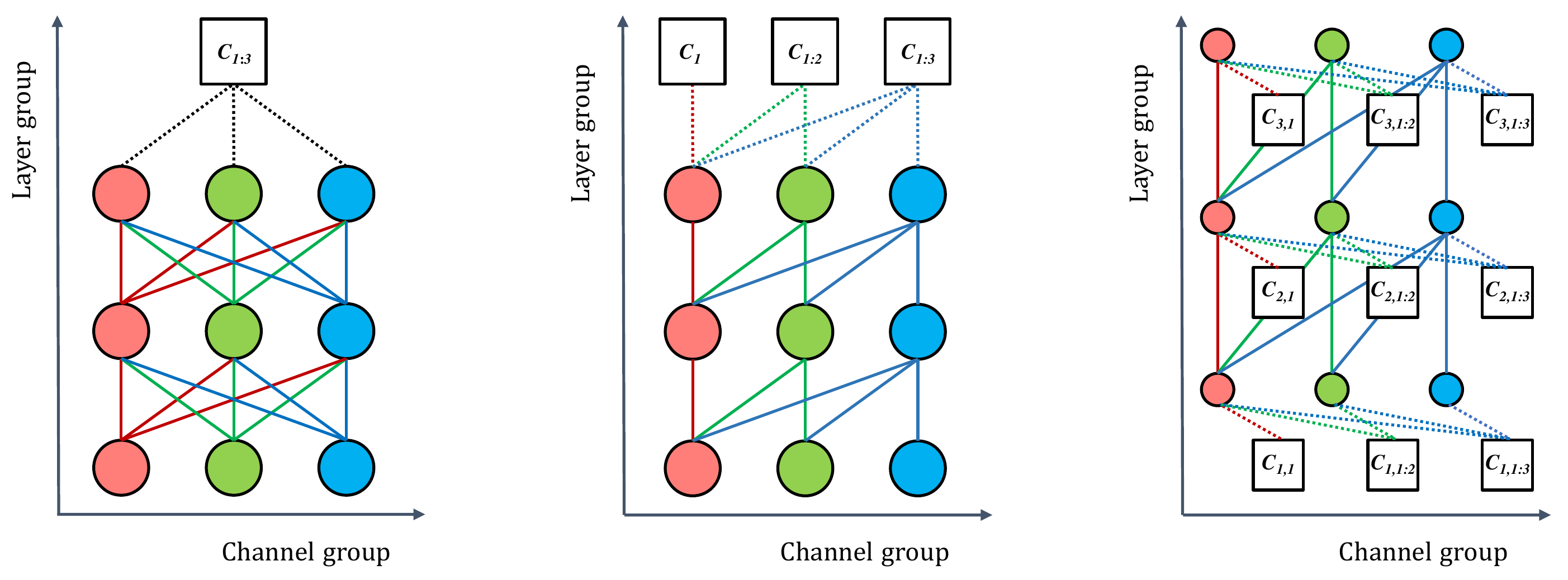}
		\caption{	Comparison of model topology with respect to slice availability. 
			(left) Conventional network in which  the channels of each feature map are fully connected to the channels of the feature maps of neighboring layers. 
			(center) Proposed depth-wise sliceable network including channel-causal convolutions  and auxiliary classifiers for channel groups.
			(right) Proposed doubly nested network with channel conditional classifiers per layer that can be sliced by channel-wise as well as layer-wise.}
		\label{fig:evolution}
	\end{subfigure}
	\begin{subfigure}[c]{\linewidth}
		\centering
		\includegraphics[width=0.95\linewidth]{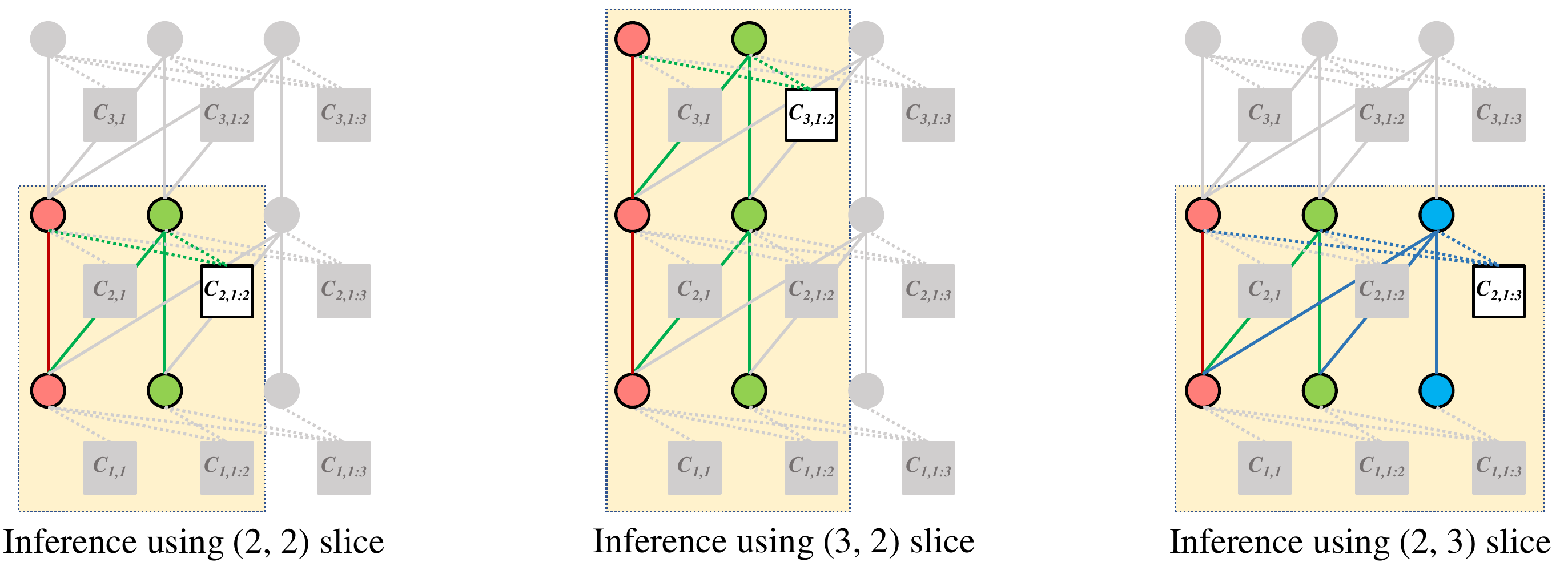}
		\caption{
        Examples of slicing according to the computational and memory demands. $(D, W)$ slice refers to a sub-network including $D$ layer groups and $W$ channel groups.}
		\label{fig:slicing_example}
	\end{subfigure}
\caption[Short]{
	Transition from conventional network to doubly nested network and its application examples.
}
\label{fig:main_concept}
\end{figure}

% \begin{figure*}
% 	\centering
% 	\includegraphics[width=0.9\linewidth]{figures/upper_conv}
% 	%	\vspace*{-0.08in}
% 	\caption{
% 		Concept figure about  the proposed upper triangular convolution
% 		\vspace*{-0.1in}
% 	}
% 	\label{fig:upper_conv}

Our goal is to design  a single network that could be sliced and used according to a variety of specifications or budget constraints without re-training.
% in which there would be little performance degradation.
In conventional convolutional neural networks (CNNs), there are full connections with trainable weights between consecutive layers in the convolution layer as well as in the fully connected layer.
Looking closer at the convolution operation between consecutive layers, output activation values from a specific layer can be computed by aggregating all output activations from previous layers as shown in the left of Fig.~\ref{fig:main_concept} (a).
That is, a simple omission of some channels in input features of specific layer might generate unpredictable output activations, which would be propagated to the following layers continuously and end up with severe performance degradation.

To solve this problem, we suggest a new neural network architecture that has restricted synaptic connections between two consecutive layers so that sliced partial models do not depend on the output from other part not included in the sliced model any more. 
More specifically, we divide the whole network into a predefined number of groups along the channels, and design a channel-causal convolution filters so that the $i$th channel group in one layer is calculated only with activation values from the channel groups from the first to $i$th  channel group in the previous  layer. 
This specialized architecture enables a partial model only having $i$th or lower channel groups to work independently with ($i+1$)th or higher channel groups both in training and in inference processes.
In this way, the convolution filters are connected in series, and the classifiers are constructed in  a similar way.
Concretely,  the channels of the feature maps passing through the last convolution filter are conditionally connected to the classifiers with the same size as the number of channel groups, so that the $i$th classifier receives the input from the first channel to the  $i$th channel as can be seen in the center of Fig.~\ref{fig:main_concept} (a).

\begin{figure*}
	\centering
	\includegraphics[width=0.97\linewidth]{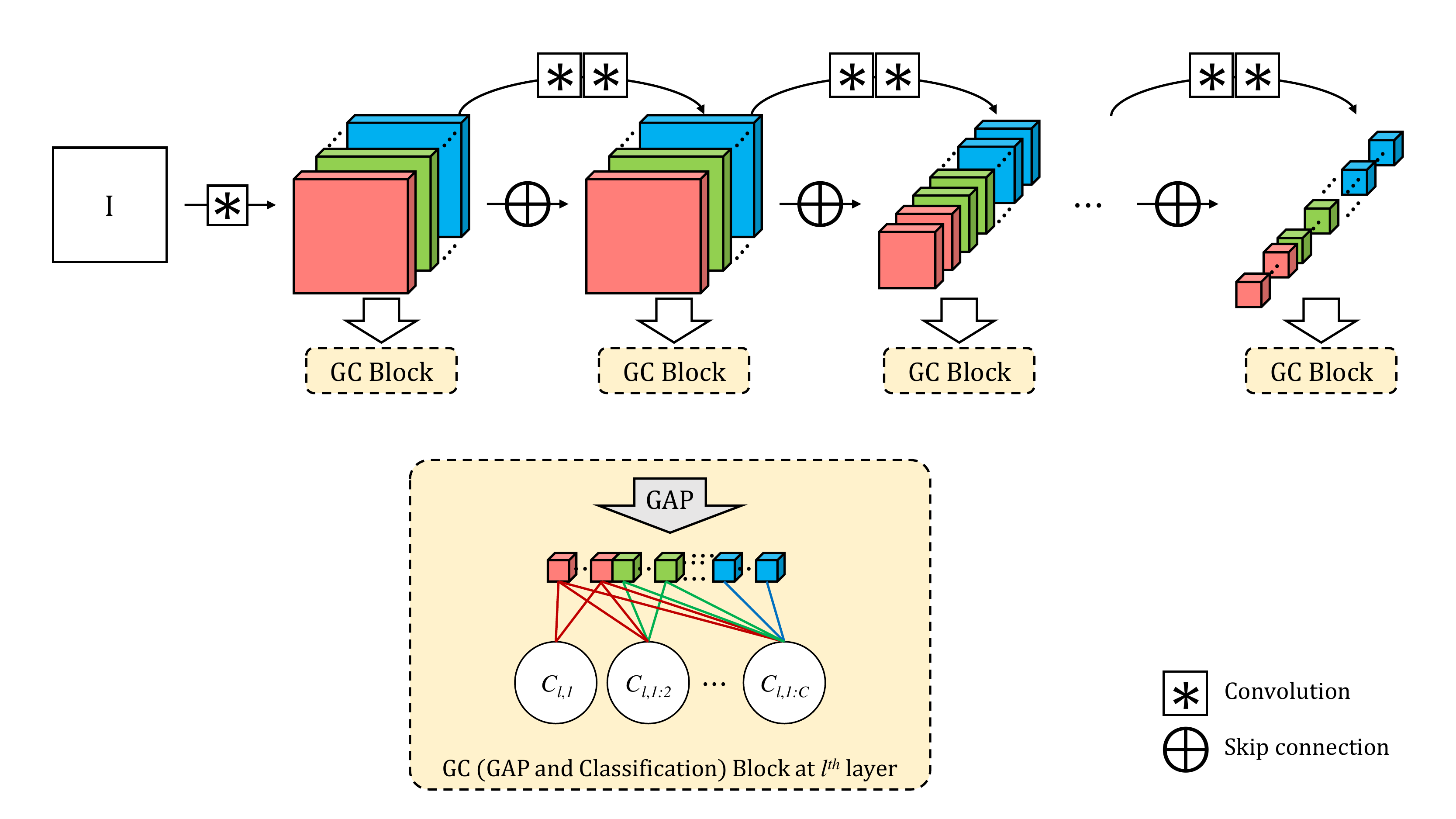}
	%	\vspace*{-0.08in}
	\caption{
		Concept illustration of a proposed architecture based on the ResNet model. Classifier is denoted by a circle in which a left part of the comma indicates covered layers and a right part is covered channel groups.  Global average pooling, intermediate channel conditional classifiers in each layer enable  a layer and channel-wise sliceable architecture.
		\vspace*{-0.1in}
	}
	\label{fig:overallflow}
	
\end{figure*}

Our final goal is to propose a structure that can be sliced by layer as well as channel wise.
To this end, we add intermediate classifiers to consecutive layers in the network so that learned feature maps from the previous layer can be reused in subsequent layers, which lead to a sliceable architecture along layer axis.
As a result, we assign classifiers to each layer (each residual block is used in ResNet case) in which the classifier for each layer is composed of the above-mentioned channel conditional manner as can be seen in the right of  Fig.~\ref{fig:main_concept} (a).
%Our main architecture  is based on ResNet and each classifier is attached to each residual block in our experiments.
Multi-scale schemes or dense connections between layers, which can guarantee high-performance but have relatively  high budget (e.g., memory or inference speed)  can  be considered for  layer-wise slicing.
However, the focus of this paper is to propose a sliceable architecture and ultimately to identify the relationship between the two slice criteria (i.e., layer and channel), not the performance enhancement itself.
Therefore, complementing the proposed architecture with more elaborate modules could be a future work.
Finally, we can obtain a architecture that can  be sliced in both axes (i.e., layer and channel), with a total of $L$ $\times$ $C$ classifiers through a combination of $L$ layers and $C$ channel groups as illustrated in Fig.~\ref{fig:overallflow}.
As can be seen in the figure,  all convolved feature maps of each layer group pass through the global average pooling, which outputs a single vector whose number of elements is same as the number of channels of the feature map. 

\subsection{Training 2D sliceable networks}

The  $L$ $\times$ $C$  classifiers in the training phase make it possible to generate  $L$ $\times$ $C$  partial models that can be sliced from the original architecture.
Although our model  contains $L$ $\times$ $C$ partial models with their own computational and memory requirements, we train all partial models simultaneously using a standard training technique rather than taking care of each partial model respectively with its own loss function. 
It is enabled by adopting multiple classifier nodes whose number is equal to the number of all possible partial models and a single aggregate loss function which encompasses all loss values from these classifiers.

\paragraph{Loss function details}

%In the training phase, we attach classifiers to all possible output nodes in the network, which enables to produce all possible predictions and get all loss values from them. 
The loss of the overall architecture is the sum of the losses of the partial models and we first  describe the loss of partial models on the channel axis. While our proposed network generally works with any learning problem such as classification or regression as long as typical deep neural network is applicable. For the sake of simplicity, we explain the setting of supervised classification and generalizing to other settings naturally follows.

Suppose that the size of last feature map after global average pooling is $C$ and the total number of class label is $N$.
In conventional  CNNs, logits are calculated  by multiplying the feature map $f \in \mathbb{R}^{C}$ by a weight matrix $W \in \mathbb{R}^{N \times C}$, which results in  $z_{n} = \sum_{c=1}^{C} W_{n, c}\cdot f_{c}$.
We then obtain the predicted probability distributions of class through softmax function. Finally, we can train the  network by  minimizing  loss function as follows:

\begin{equation}
\mathcal{L} =   -\frac{1}{N}\sum_{i=1}^{N}[y_{i}\:log(\hat{y}_{i})], 
\end{equation}
where ${y}$ is a ground truth, and $\hat{y}=softmax(z)$.

To construct a conditional classifier on the channel in a cumulative manner, unlike the existing techniques, we calculate the logit   in the following manner, i.e.,  $Z_{n, c} = \sum_{k=1}^{c} W_{n, k} \cdot  f_{k}$. 
This allows $Z_{\: \cdot, c}$ to use only the features from $f_{1:c}$.
We then pass the logits through softmax function, $\hat{y}_{\cdot, c} = softmax(Z_{\: \cdot, c})\:$. Finally, we can calculate the loss function in the conditional channel as:
  \begin{equation}
\mathcal{L}_{c} =   -\frac{1}{N}\sum_{i=1}^{N}[y_{i}\:log(\hat{y}_{i, c})]
 \label{eq:loss_channel}
  \end{equation}
  
Since we want to obtain an architecture that is a layer-wise slice as well as a channel-wise slice, we attach intermediate classifiers on all layers  with global average pooling.
This allows us to get the feature map $ f^l $ for each layer.
If the channel conditional classifier is applied to each layer similarly to  (\ref{eq:loss_channel}), we can obtain loss function for the architecture which can be sliced layer-wise as well as channel-wise as follows:
  \begin{equation}
\mathcal{L}_{c}^l =  -\frac{1}{N}\sum_{i=1}^{N}[y_{i}\:log(\hat{y}_{i, c}^l)], 
\label{eq:loss_total}
\forall (l, c) \in \{1, 2, ..., L\} \times \{1, 2, ..., C\},
\end{equation}
where $\hat{y}_{i, c}^l$ is softmax output of $Z_{n, c}^l = \sum_{k=1}^{c} W_{n, k}^l \cdot  f_{k}^l$ of the $i$th sample.

% For each layers, apply global average pooling, we get input of layer classifier $x \in \mathbb{R}^{F} $ and classifier weights $W \in \mathbb{R}^{F \times C}$ then output logits $z_{c} = \sum_{i=1}^{F} W_{i,c}x_{i}$ and we can get probability from logits apply softmax $p(y=c|x) = \frac{e^{z_{c}}}{\sum_{i=1}^{C} e^{z_{i}}}$. but, for channel slicing, We want to get sliced probability $p(y=c|x_{1:f})$ for all $f \in \{1, 2, ... , F\}$. We can model this without any additional parameter, just sharing parameter for C classifiers, $Z_{f, c} = \sum_{i=1}^{f} W_{i, c}x_{i}$ and use $p(y=c|x_{1:f})=\frac{e^{Z_{f, c}}}{\sum_{i=1}^{C} e^{Z_{f, i}}}$.
% For computational efficiency, we can compute $Z_{f,c} = Z_{f-1,c} + W_{f,c}x_{f}$

% For all layer $l \in \{1, 2, ..., L\}$

% However, we do not need all classifier nodes if we need only some of partial models when using the selected models in the inference phase. We can easily cut off unnecessary parameters from the whole classifiers' parameters and the resulting operations also as we are doing for conventional convolutional layers. 

% \begin{figure*}[htbp]
% \centering
% \begin{subfigure}{0.63\textwidth}
%   \includegraphics[height=5.5cm]{figures/classifier_details.png}
%   \caption{}
% \end{subfigure}
% \caption{Classifier stage}
% \label{fig:classifier_details}
% \end{figure*}

\paragraph{An aggregate loss function}\label{secttion:loss_func}

An individual loss for each partial model can be calculated using the obtained logit values and the target value by a conventional loss function, which is a cross-entropy in our case. We make a single objective function by combining all loss functions rather than optimizing each loss function respectively. Our baseline implementation of this aggregate objective function is simply a sum of all loss functions coming from all possible partial models. Applying standard backpropagation to this single loss function enables its gradient values to flow into all connected loss functions and optimize all partial models at the same time for each iteration. 

\begin{equation}
\mathcal{L} = \frac{\sum_{l, c}\lambda(l, c) \mathcal{L}_{c}^{l}}{\sum_{l, c}\lambda(l, c)} 
\end{equation}

We extend this baseline aggregate loss function by introducing additional parameters each of which serves as a multiplier to the individual loss function, which is a weight to the loss function, and reflects users' preferences. 
These parameters can be represented by a single matrix whose individual element indicates a relative importance of the corresponding partial models in terms of the final performance (e.g., classification accuracy). 

In this study, we propose three weight matrices. 
The first type ($\lambda_{descend}(l, c) = \gamma^{-(l+c)}$) focuses on low-complexity models. The weight ($\lambda(l, c)$) to the loss function of a specific model with $l$ layer groups and $c$ channel groups decays exponentially as $l$ or $c$ gets larger, where $\gamma$ is a hyper-parameter larger than one. On the contrary, the second type ($\lambda_{ascend}(l, c) = \gamma^{(l+c)}$) focuses on high-performance partial models. In addition we  can customize the loss weight matrix, i.e., $\lambda_{custom}(l, c) = f(l, c)$.
We verified this prioritization scheme works actually as we intended and the quantitative results and other details are shown in Section \ref{section:experiments}.

% \section{Experiments}
% \subsection{\textcolor{green}{Xhark} 데이터셋 및 네트워크 구조 detail}
% (MNIST, SVHN, Cifar10) (TBD) 데이터셋을 사용하여 학습하였다. 네트워크는 ResNet을 사용하였다.
% \subsection{\textcolor{green}{Xhark} Width and/or Depth 상관관계 실험}
% Flexible하게 width, depth를 조절할 수 있는 ResNet을 통해 width와 depth의 상관관계를 실험하였다.
%   \begin{figure*}[htbp]
%   \centering
%   \begin{subfigure}{1.00\textwidth}
%     \includegraphics[height=5.5cm]{figures/layer_channel_acc.png}
%     \caption{}
%   \end{subfigure}
%   \caption{Layer Channel Acc}
%   \label{fig:layer_channel_acc}
%   \end{figure*}
  
%   \begin{figure*}[htbp]
%   \centering
%   \begin{subfigure}{1.00\textwidth}
%     \includegraphics[height=5.5cm]{figures/stats.png}
%     \caption{}
%   \end{subfigure}
%   \caption{Controllable range}
%   \label{fig:stats}
%   \end{figure*}

% \subsection{\textcolor{green}{Xhark} ablation study}
% \subsubsection{}

\section{Experiments}\label{section:experiments}

\subsection{Experimental setup}

We evaluated the proposed method in widely used CIFAR-10. We also used the street view house numbers (SVHN) dataset.
% We evaluated the proposed method in widely used CIFAR-10 containing  50,000 training images and 10,000 test images where each image is 32x32 pixel.
% Among training images, 5,000 images are used as a validation set.
% We also used the street view house numbers (SVHN) dataset that  contains 73257 digits for training and 26032 digits for testing of size 32x32.
% Following the work in \cite{DBLP:conf/cvpr/HeZRS16}, we apply conventional data augmentation to the training images. Concretely, we zero-padded each image with four pixels and then randomly cropped 32×32 images. All images were horizontally flipped with a probability of 0.5 and normalized using channel means and standard deviation.
Our architecture is based on the ResNet-32 model, which consists 15 residual blocks and one fully-connected layer.
Our architecture includes a classifier between the residual blocks, and the first feature map has 16 channels, so it contains a total of 256 (16x16) classifiers.
To train this architecture, we use a momentum optimizer with a 0.9 momentum term with mini-batch 
size 128. 
Our models are trained for 40k steps, with an initial learning rate of 0.1, which is divided by a factor 10 after 60k steps and 60k steps.

\subsection{Performance change according to channel-wise slice}

\begin{figure*}[htbp]
	\centering
	\begin{subfigure}[b]{0.49\textwidth}
		\includegraphics[height=3.8cm]{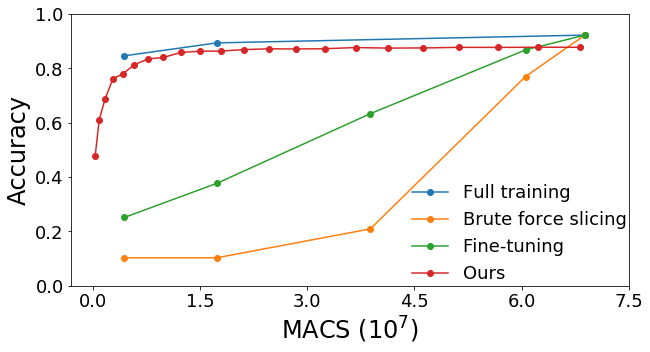}
		\caption{CIFAR-10}
	\end{subfigure} 
	\begin{subfigure}[b]{0.49\textwidth}
		\includegraphics[height=3.8cm]{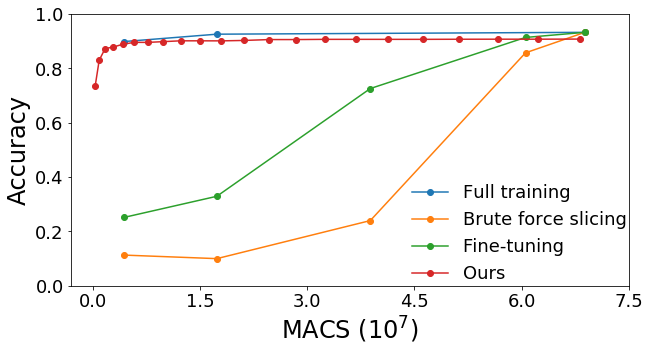}
		\caption{SVHN}
	\end{subfigure}
	\caption{
		Classification accuracy with respect to the number of sliced channels
		\vspace*{-0.1in}
	}
	\label{fig:width_control}
\end{figure*}

One of the key advantages of the proposed architecture is that it is possible to slice on the channel axis, with less performance degradation.
Fig.~\ref{fig:width_control} shows the performance changes of the proposed scheme and comparison models when sliced by channel on CIFAR-10 and SVHN datasets.
The proposed architecture is based on Resnet-32 model. 
Therefore, the lower bound of the performance could be a classification accuracy obtained by the scenario of truncating the channel in the existing Resnet model.
As expected, conventional CNNs has serious performance degradation by the  channel slicing because  there are full connections between consecutive layers.
On the other hand, the upper bound of performance is the case of fine-tuning the entire network again after truncating channels in the original network.
In addition, a model in which the parameters of all channels are fixed and only the classifier is re-trained after cutting the channel from the original network.
Our model shows relatively low performance degradation even though the total channel is getting smaller in both datasets.

\begin{figure*}[htbp]
	\centering
% 	\begin{subfigure}{0.8\textwidth}
	\begin{subfigure}{0.6\textwidth}
		\includegraphics[height=4.0cm]{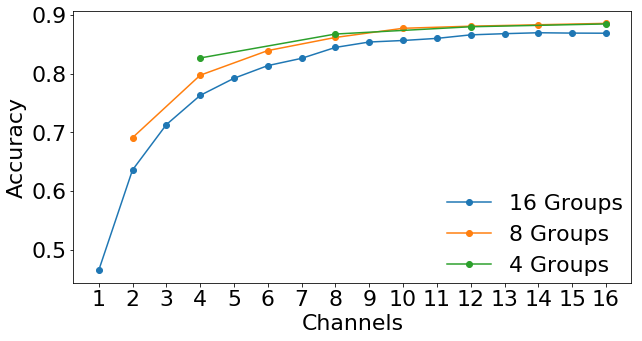}
	\end{subfigure}
	\caption{
		Performance change with respect to channel-wise slicing  when adjacent channels are grouped.
		\vspace*{-0.1in}
	}
	\label{fig:slice_level_graph}
\end{figure*}

When constructing a sliceable architecture based on the channel, our base model handles all the channels separately. 
% For example, since the first feature map of the proposed architecture has a total of 16 channel sizes, a total of 16 slices can be constructed on the channel axis.
During  training our  sliceable architecture, instead of processing all sliced channels separately, adjacent channels can be  processed by group.
In this experiment, we compared the performance changes with respect to channel slicing when learning the architecture with 8 slices (i.e., processing two consecutive channels together) and 4 slices (i.e., processing 4 consecutive channels together) in addition to base model with 16 slices.
As can be seen in Fig.~\ref{fig:slice_level_graph}, we can see a tradeoff between performance degradation and sliceable degrees of freedom when grouping channels.

%% \subsection{Performance change according to slice layer}

%% TODO_FIXME : Supplementary Materials

%% The proposed architecture can slice on the layer axis as well as on the channel.
%% In this experiment we investigated the effect of slicing on the slice axis when the channel size was fixed.
%% Experiments were conducted by dividing into 16 channels without slicing, half channel, and 1/4 slice.
%% The proposed architecture consists of a total of 16 residual blocks, and we can slice 16 times because we attach a classifier to each residual block.
%% In this experiment, residual blocks and layers are used  interchangeably.
%% From Fig.~\ref{fig:layer_wise_exp}, we ​​can see that the more layers sliced ​​from the base architecture, the lower the performance

%Interestingly, there is a significant performance change between the 6th and 7th layers and between the 11th and 12th layers.

%%\begin{figure*}[htbp]
%%	\centering
% 	\begin{subfigure}{0.8\textwidth}
%%	\begin{subfigure}{0.6\textwidth}
%%		\includegraphics[height=4.0cm]{figures/layer_wise_exp_v2.png}
%%	\end{subfigure}
%%	\caption{
%%	Performance changes when channel is divided into three cases and sliced to layer axis
%%		\vspace*{-0.1in}
%%	}
%%	\label{fig:layer_wise_exp}
%%\end{figure*}

\subsection{Effects of slicing both in width and depth directions}
\begin{figure}[ht!]
	\centering
% 	\begin{subfigure}{0.3\textwidth}
% % 		\includegraphics[height=5.5cm]{figures/16x16_accuracies.png}
% 		\includegraphics[height=5cm]{figures/16x22_accuracies_3d.png}
% % 		\caption{Classification accuracies}
% 	\end{subfigure}
% 	\begin{subfigure}{0.3\textwidth}
% % 		\includegraphics[height=5.5cm]{figures/16x16_accuracies.png}
% % 		\includegraphics[height=5.5cm]{figures/16x16_accuracies_2d.png}
% 		\includegraphics[height=5cm]{figures/16x22_accuracies_2d.png}
% 	\end{subfigure}
% % 	\includegraphics[height=7cm]{figures/16x16_accuracies_2d.png}
% % 	\caption{
% %     Classification accuracies of 16$\times$16 sliced models with the baseline loss function
% % %     \vspace*{-0.1in}
% %     }
	\includegraphics[height=4cm]{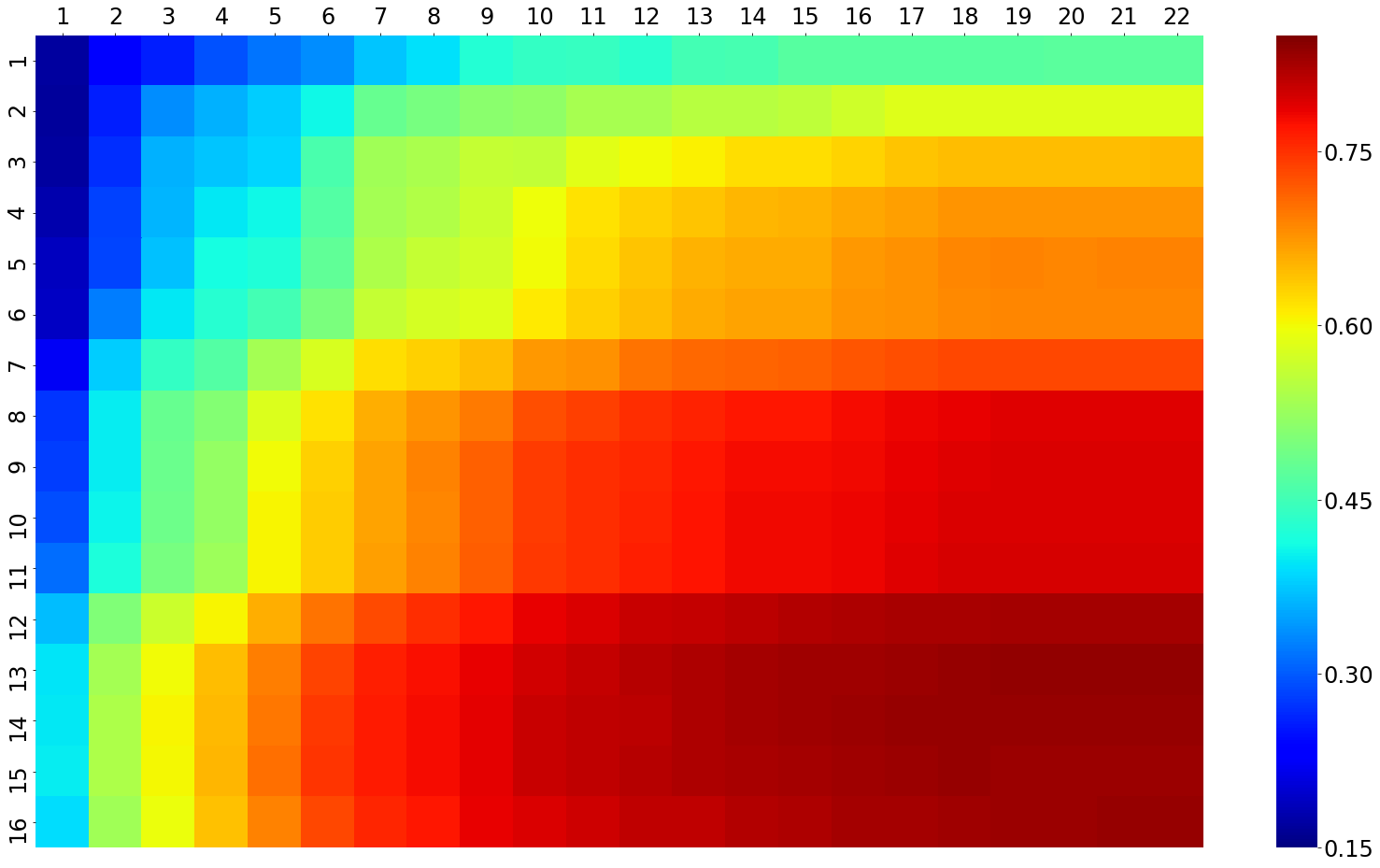}
    \includegraphics[height=4cm]{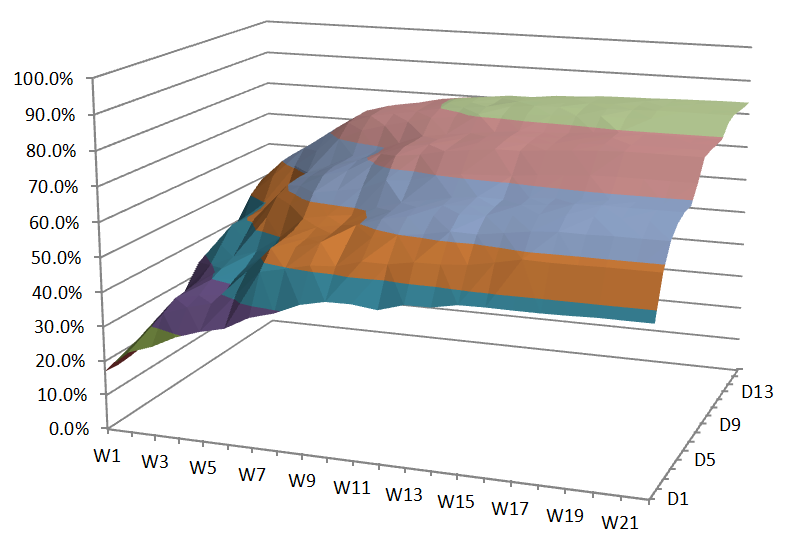}
    \caption{
    Classification accuracies with 16$\times$22 slicing
    (D$i$: up to $i$-th layer groups, W$j$: up to $j$-th channels groups)}
	\label{fig:16x22_accuracies}
% \end{figure*}
\end{figure}

\paragraph{Slicing with the baseline loss function}
Fig.~\ref{fig:16x22_accuracies} shows classification accuracies of all partial models when training the full network with up to 16 layer groups and 22 channel groups, which is the finest-grained grouping in our setup. As intended, the wider and deeper neural network models show the higher accuracies without degrading the performance of the largest model seriously.

% \begin{figure*}[htbp]

\paragraph{Effects of non-flat loss weight matrices}
In Section \ref{secttion:loss_func}, we have introduced the loss weight matrix $\lambda(i, j)$ to prioritize the constituent partial models. Other than the flat matrix used in the baseline loss function, we explore three variants of loss weight matrices. 

\begin{figure}[ht!]
    \centering
    \begin{subfigure}[b]{0.32\textwidth}
        \includegraphics[width=\textwidth]{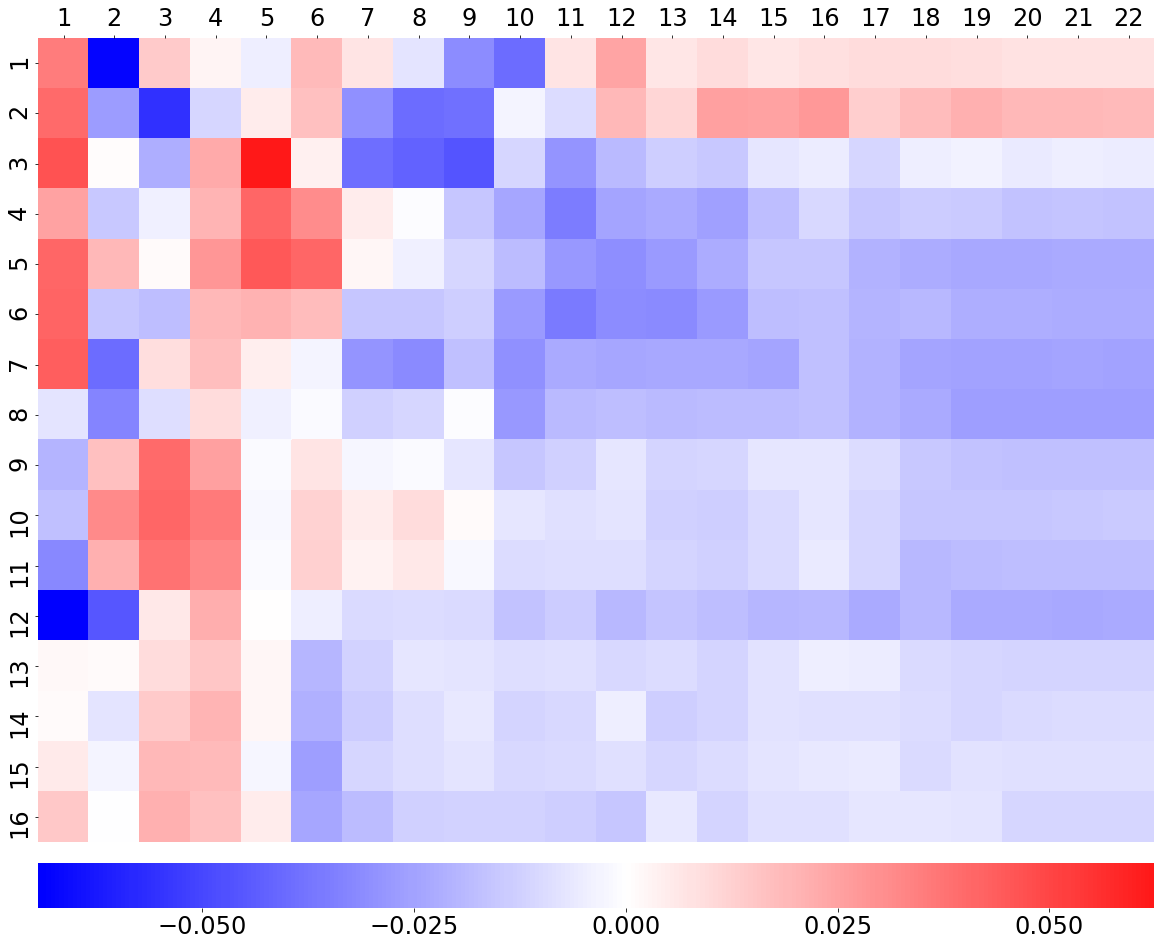}
        \caption{Low cost-preferred}
        \label{fig:16x22_dec_w_diff}
    \end{subfigure}    
    \begin{subfigure}[b]{0.32\textwidth}
        \includegraphics[width=\textwidth]{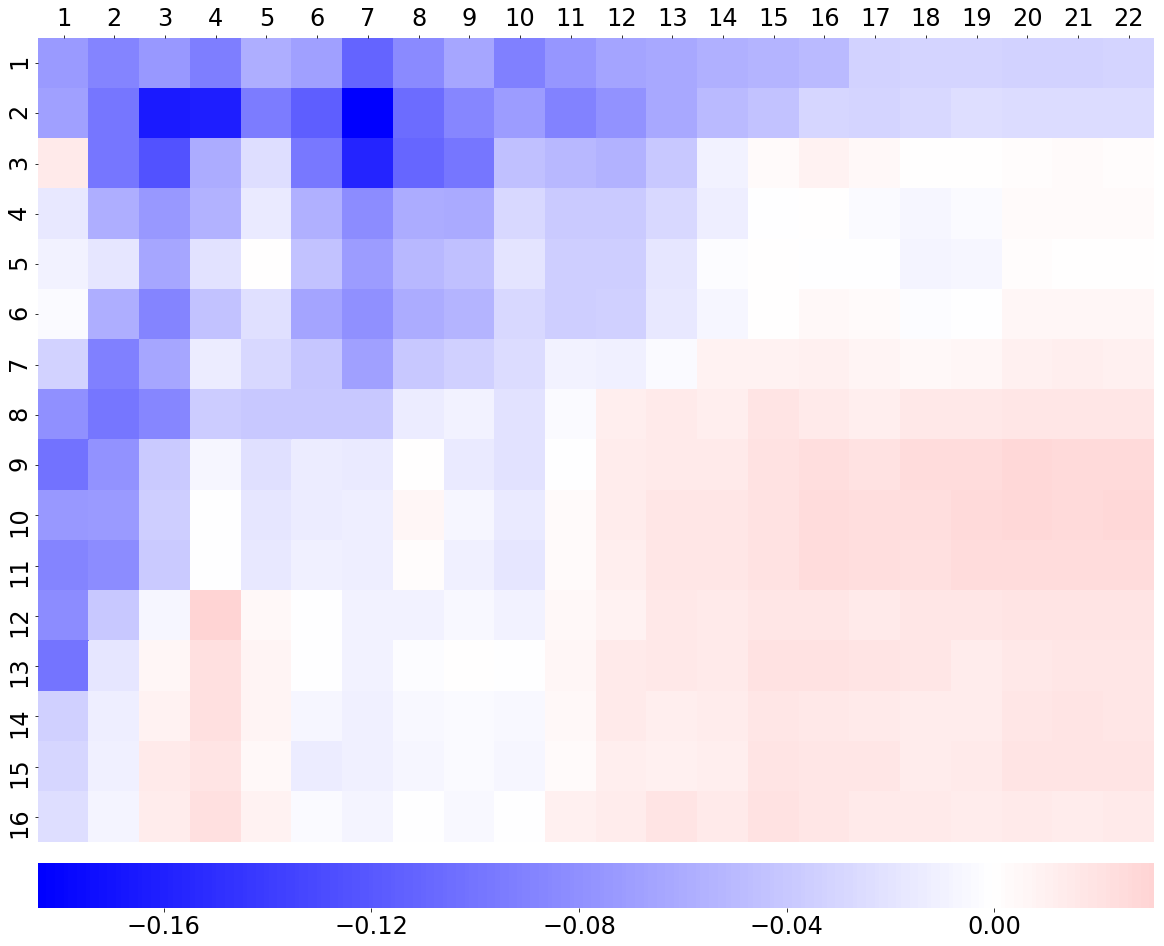}
        \caption{High performance-preferred}
        \label{fig:16x22_inc_w_diff}
    \end{subfigure}
    \begin{subfigure}[b]{0.32\textwidth}
        \includegraphics[width=\textwidth]{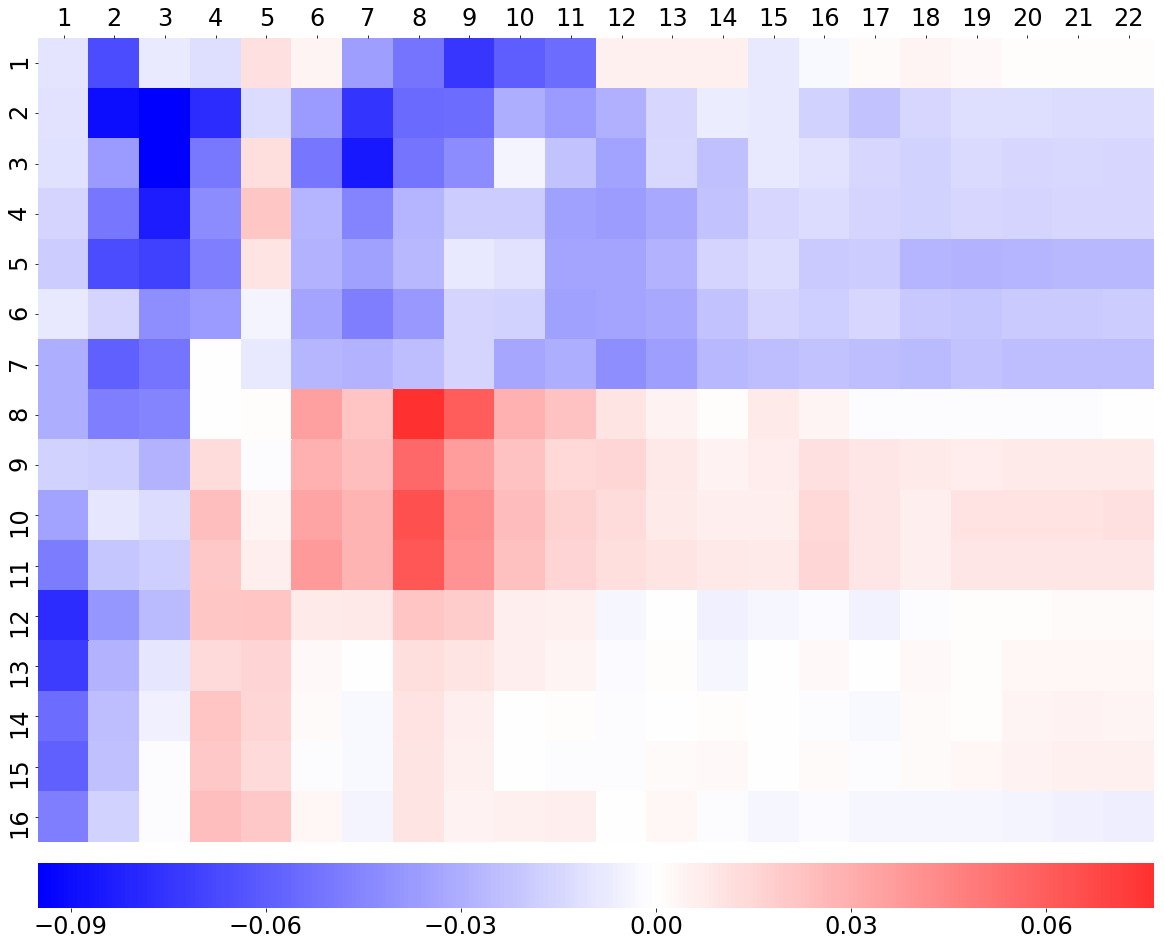}
        \caption{100-times weight to (L8, C8)}
        \label{fig:16x22_custom_pick_w_diff}
    \end{subfigure}
    \caption{Accuracy differences from the baseline for various loss weight matrices
    (red: higher than the baseline, blue: lower than the baseline)}
    \label{fig:loss_weight_plot}
\end{figure}

Fig.~\ref{fig:16x22_dec_w_diff} and ~\ref{fig:16x22_inc_w_diff} show the changes of classification accuracies from the baseline when we apply two different loss weight matrices that give more weights to low cost models and low performance models respectively. The results show that the partial models with higher weights are able to achieve higher performance than the baseline while penalizing other models with relatively lower weights as expected. We can observe that the impact is more distinguishable in the low cost-preferred training also. 

We also tested a case to give a high loss weight to a single specific model. When we assign a 100-times weight only to the model (L8, C8) in the center ($\lambda(i, j)=100$ for i=8, j=8, otherwise $\lambda(i, j)=0$), the result in Fig.~\ref{fig:16x22_custom_pick_w_diff} shows that the target model outperforms the baseline with the same configuration by 7.2\%.

We present another example to show the advantages of the doubly nested network over the previous slicing scheme only with a single degree of freedom in the supplementary material.

\section{Discussion}

This paper proposes a neural network architecture called DNNet in which all models share parameters maximally by nesting all-in-one.
This nested structure allows for slicing into channel wise or layer wise while maintaining its original performance as high as possible.
For channel-wise slice which has not been explored to date, we design channel-causal convolutions which  sort channels topologically and connecting neurons accordingly.
In addition, we add intermediate classifiers to consecutive layers in the network so that learned feature maps from the previous layer can be reused in subsequent layers, which lead to a sliceable architecture along layer axis.

Through various experiments, we show that the doubly nested network is robust to the channel-level slicing without causing severe performance degradation. We also verify that the channel-level slicing can be integrated successfully with the layer-level slicing so that we can benefit from increased degrees of freedom while leading to better solutions in terms of computational efficiency. 

\subsubsection*{Acknowledgments}
The authors would like to thank Jung Kwon Lee for the valuable feedback on this work and Nam-Gyu Cho for the great illustration of whole concept as figures.
 % Do not include acknowledgments in the anonymized submission

% \input{text/References.tex}

\bibliographystyle{abbrv}
\bibliography{References}

\end{document}